\documentclass[journal]{IEEEtran}

\ifCLASSINFOpdf
   \usepackage[pdftex]{graphicx}
   \graphicspath{{..f/}{../jpeg/}}
   \DeclareGraphicsExtensions{.pdf,.jpg,.png}
\else
   \usepackage[dvips]{graphicx}
   \graphicspath{{../eps/}}
   \DeclareGraphicsExtensions{.eps}
\fi

\usepackage{amsmath,amssymb,amsthm}
\usepackage{fancyhdr}
\usepackage[ruled,linesnumbered]{algorithm2e}

\renewcommand{\mathbf}{\boldsymbol}

\begin{document}

\title{Improving training of deep neural networks via Singular Value Bounding}

\author{Kui~Jia
\\
School of Electronic and Information Engineering \\
South China University of Technology \\
\texttt{kuijia@scut.edu.cn}

}

%

\maketitle

\begin{abstract}

Deep learning methods achieve great success recently on many computer vision problems, with image classification and object detection as the prominent examples. In spite of these practical successes, optimization of deep networks remains an active topic in deep learning research. In this work, we focus on investigation of the network solution properties that can potentially lead to good performance. Our research is inspired by theoretical and empirical results that use orthogonal matrices to initialize networks, but we are interested in investigating how orthogonal weight matrices perform when network training converges. To this end, we propose to constrain the solutions of weight matrices in the orthogonal feasible set during the whole process of network training, and achieve this by a simple yet effective method called Singular Value Bounding (SVB). In SVB, all singular values of each weight matrix are simply bounded in a narrow band around the value of $1$. Based on the same motivation, we also propose Bounded Batch Normalization (BBN), which improves Batch Normalization by removing its potential risk of ill-conditioned layer transform. We present both theoretical and empirical results to justify our proposed methods. Experiments on benchmark image classification datasets show the efficacy of our proposed SVB and BBN. In particular, we achieve the state-of-the-art results of $3.06\%$ error rate on CIFAR10 and $16.90\%$ on CIFAR100, using off-the-shelf network architectures (Wide ResNets). Our preliminary results on ImageNet also show the promise in large-scale learning.

\end{abstract}



\IEEEpeerreviewmaketitle

\section{Introduction}
\label{SecIntro}

Deep learning methods keep setting the new state-of-the-art for many computer vision problems, with image classification \cite{ILSVRC15} and object detection \cite{CoCoECCV} as the prominent examples. These practical successes are largely achieved by newly proposed deep architectures that have huge model capacities, including the general ones such as Inception \cite{InceptionResNet} and ResNet \cite{ResNet,WideResNet}, and also specially designed ones such as Faster R-CNN \cite{FasterRCNN} and FCN \cite{FCN}. Training of these ultra-deep/ultra-wide networks are enabled by modern techniques such as Batch Normalization (BN) \cite{BatchNorm} and residual learning \cite{ResNet}.

In spite of these practical successes, however, optimization of deep networks remains an active topic in deep learning research. Until recently, deep networks are considered to be difficult to train. Researchers argue for different reasons causing such difficulties, such as the problem of vanishing/exploding gradients \cite{XavierInit,RNNTrainDifficulty}, internal shift of feature statistics \cite{BatchNorm}, and also the proliferation of saddle points \cite{TrustRegion4SaddlePoint,DeepLearningNoPoorLocalMinima}. To address these issues, different schemes of parameter initialization \cite{XavierInit,ExactSolution}, shortcut connections \cite{ResNet,PreActResNet}, normalization of internal activations \cite{BatchNorm}, and second-order optimization methods \cite{TrustRegion4SaddlePoint} are respectively proposed.

In this work, we focus on another important issue to address the difficulty of training deep neural networks. In particular, given the high-dimensional solution space of deep networks, it is unclear on the properties of the (arguably) optimal solutions that can give good performance at inference. Without knowing this, training by a specified objective function easily goes to unexpected results, partially due to the proliferation of local optima/critical points \cite{TrustRegion4SaddlePoint,DeepLearningNoPoorLocalMinima}. For example, it is empirically observed in \cite{ResNet} that adding extra layers to a standard convolutional network (ConvNet) does not necessarily give better image classification results. This unclear issue is further compounded by other (aforementioned) optimization difficulties.

Existing deep learning research has some favors on the solutions of network parameters, and also on network architectures that can give desirable solutions. In particular, Arpit \emph{et al.} \cite{OptCon4NetSparseRecovery} study the properties of network parameters that can ensure accurate recovery of the true signals of hidden representations, and prove that for sparse true signals, e.g., those out of Rectified Linear Units (ReLU) activations \cite{ReLU}, strong recovery can be achieved if the weight matrix is highly incoherent. Saxe \emph{et al.} \cite{ExactSolution} advocate orthogonal initialization of weight matrices, and theoretically analyze its effects on learning efficiency using deep linear networks. Practical results on image classification using orthogonal initialization are also presented in \cite{GoodInit}. In terms of the favored properties on network architectures, the development of Inception models \cite{InceptionResNet} relies on the Hebbian principles \cite{ProvableBoundHebbian}, and ResNet \cite{ResNet} argues for residual learning by shortcut connections.

In this paper, we are inspired by the analysis of orthogonal initialization in \cite{ExactSolution}, and aim to constrain the solutions of weight matrices in the orthogonal feasible set during the whole process of network training. To this end, we propose a simple yet effective method called Singular Value Bounding (SVB). In SVB, all singular values of each weight matrix are simply bounded in a narrow band around the value of $1$ (Section \ref{SecAlgorithm}). When using stochastic gradient descent (SGD) or its variants for network training, this amounts to turning SVB on by every a specified number of iterations. We present theoretical analysis, using deep linear networks, to show how such learned networks are better on forward-propagation to achieve training objectives, and backward-propagation of training errors (Section \ref{SecTechAnalysis}).



Batch normalization \cite{BatchNorm} is a very effective method to improve and accelerate network training. We prove that in the framework of our theoretical analysis, trainable parameters in BN may cause ill-conditioned layer transform. We thus propose Bounded Batch Normalization (BBN), a technique that improves BN by removing this risk without sacrificing all its other benefits. BBN achieves this by simply bounding the values of BN parameters during training.

We present benchmark image classification experiments using both ConvNets \cite{VGGNet} and modern network architectures \cite{PreActResNet,WideResNet,InceptionResNet} (Section \ref{SecExps}). Our results show that SVB indeed improves over SGD based methods for training various architectures of deep networks, and in many cases with a large margin. Our proposed BBN further improves over BN. In particular, we achieve the state-of-the-art results of $3.06\%$ error rate on CIFAR10 and $16.90\%$ on CIFAR100 \cite{Cifar}, using off-the-shelf network architectures (Wide ResNets \cite{WideResNet}). Our preliminary results on the large-scale ImageNet dataset are consistent with those on moderate-scale ones.

\section{Related works}
\label{SecLitecture}

In this section, we briefly review the closely related deep learning methods that also pay attention to the properties of network solutions.

Saxe \emph{et al.} \cite{ExactSolution} theoretically study the gradient descent learning dynamics of deep linear networks, and give similar empirical insights for deep nonlinear networks. They further suggest that using orthogonal initialization of weight matrices can achieve learning efficiency similar to that of unsupervised pre-training. Mishkin and Matas \cite{GoodInit} present promising results on image classification, using the orthogonal initialization idea in \cite{ExactSolution}. Our theoretical analysis in Section \ref{SecTechAnalysis} follows \cite{ExactSolution}, but are different in the following aspects. We focus on studying the conditions when network training converges, while \cite{ExactSolution} focuses on the conditions right after network initialization. Our analysis centers around our proposed SVB method, and we discuss how SVB can resolve the issues that appear as the network training proceeds. We also extend our theoretical analysis to BN \cite{BatchNorm}, and propose a new BBN method that improves over BN for training modern deep networks.


Arpit \emph{et al.} \cite{OptCon4NetSparseRecovery} also study the properties of network parameters that can have good performance, but from a signal recovery point of view. In particular, they study the reverse data-generating properties of auto-encoders where input samples are generated from the true signals of hidden representations. They prove that for sparse true signals, e.g., those out of ReLU activations, strong recovery can be achieved if the weight matrix is highly incoherent. Different from \cite{OptCon4NetSparseRecovery}, our main concern is on the properties of network parameters that can give good image classification performance by feed-forward computations.

In \cite{RNNTrainDifficulty}, a soft constraint technique is proposed to deal with the vanishing gradient in training recurrent neural networks (RNNs). The soft constraint regularizes the learning of weight matrices so that those better to achieve norm preservation of error signals across layers are favored. In contrast, our proposed SVB method directly controls the singular values of weight matrices, and norm preservation of error signals is only part of our benefits.

A recent work from Wisdom and Powers \emph{et al.} \cite{FullCapacityUnitaryRNN} shows full-capacity unitary recurrence matrices can be used in RNNs, and can be optimized over the differentiable manifold of unitary matrices. This improves over \cite{UnitaryRNN}, where unitary recurrent matrices are restricted to be a product of parameterized unitary matrices. In contrast, we focus on convolutional networks in this work, where weight matrices are not square. We only enforce column or row vectors of weight matrices of ConvNets to be near orthogonal, while giving them more flexibility to better learn to the training tasks. This relaxation from strict orthogonality enables us to use very simple algorithms compatible with standard SGD based training. Practicably, we observe our SVB algorithm is just as efficient as SGD based training, while achieving the property of near orthogonality.

Other very recent relevant research includes \cite{OzayOkataniCNNKernelSubManifold} that develops geometric framework to analyze network optimization on sub-manifolds of certain normalized kernels, including orthonormal weight matrices, and proposes a SGD based algorithm to optimize on the sub-manifolds with guaranteed convergence. In \cite{EDJM}, Wang \emph{et al.} propose Extended Data Jacobian Matrix (EDJM) as a network analyzing tool, and study how the spectrum of EDJM affects performance of different networks of varying depths, architectures, and training methods. Based on these observations, they propose a spectral soft regularizer that encourages major singular values of EDJM to be closer to the largest one (practically implemented on weight matrix of each layer). This is related, but different from our proposed hard constraint based SVB method.

\section{The proposed Singular Value Bounding algorithm}
\label{SecAlgorithm}

Suppose we have $K$ pairs of training samples $\{\mathbf{x}_i, \mathbf{y}_i\}_{i=1}^K$, where $ \mathbf{x}_i \in \mathbb{R}^{N_x} $ is a training input and $y_i$ is its corresponding output. $\mathbf{y}_i \in \mathbb{R}^{N_y}$ could be a vector with continuous entries for regression problems, or a binary one-hot vector for classification problems. A deep neural network of $L$ layers performs cascaded computations of $ \mathbf{x}^l = f(\mathbf{z}^l) = f(\mathbf{W}^l\mathbf{x}^{l-1} + \mathbf{b}^l) \in \mathbb{R}^{N_l}$ for $l=1, \dots, L$, where $\mathbf{x}^{l-1} \in \mathbb{R}^{N_{l-1}}$ is the input feature of the $l^{th}$ layer, $f(\cdot)$ is an element-wise activation function, and $\mathbf{W}^l \in \mathbb{R}^{N_l\times N_{l-1}}$ and $\mathbf{b}^{l} \in \mathbb{R}^{N_l}$ are respectively the layer-wise weight matrix and bias vector. We have $\mathbf{x}^0 = \mathbf{x}$. With appropriate training criteria, network optimization aims to find solutions of network parameters $\Theta = \{ \mathbf{W}^l, \mathbf{b}^l\}_{l=1}^L$, so that the trained network is able to produce good estimation of $\mathbf{y}$ for any test sample $\mathbf{x}$.

Training of deep neural networks is usually based on SGD or its variants \cite{Momentum}. Given the training loss function ${\cal{L}}\left(\{\mathbf{x}_i, \mathbf{y}_i\}_{i=1}^K; \Theta\right)$, SGD updates $\Theta$ based on a simple rule of $\Theta_{t+1} \leftarrow \Theta_t - \eta \frac{\partial{\cal{L}}}{\partial{\Theta_t}}$, where $\eta$ is the learning rate. The gradient $\frac{\partial{\cal{L}}}{\partial{\Theta_t}}$ is usually computed from a mini-batch of training samples. Network training proceeds by sampling for each iteration $t$ a mini-batch from $\{\mathbf{x}_i, \mathbf{y}_i\}_{i=1}^K$, until a specified number $T$ of iterations or the training loss plateaus.

\begin{figure}[t]
\centering

\includegraphics[scale=0.45]{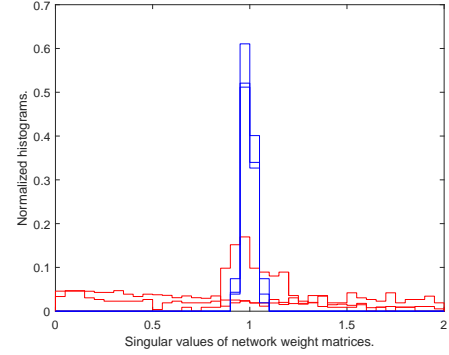}  \\

\caption{ {\small Normalized singular value histograms for weight matrices of a $38$-layer ConvNet trained for CIFAR10 image classification (cf. Section \ref{SecExpsConvNetStudies} for details of the network architecture). Red stairs of histograms are from SGD based training, and blue ones are from our proposed SVB method. For both methods, three histograms respectively for lower, middle, and higher network layers are counted when network training converges from orthogonal initializations. Given the sharp difference of singular value spectra between the two methods, it is interesting to observe that both methods give reasonably good performance, and our method even outperforms SGD based training.  }  }  \label{FigIllus}
\end{figure}

Existing deep learning research suggests that in order to get good performance, initializations of $\Theta$ matter. In particular, scaled random Gaussian matrices are proposed in \cite{XavierInit,XavierimprovedInit} as the initializations of weight matrices $\{ \mathbf{W}^l\}_{l=1}^L$, and random orthogonal ones are advocated in \cite{ExactSolution,GoodInit}. Given different initializations, these methods train deep networks using SGD or its variants. Theoretical analysis in \cite{ExactSolution} and empirical results in \cite{GoodInit} demonstrate some advantages of orthogonal initializations over Gaussian ones. In this work, we are interested in pushing a step further to know what solutions of $\Theta$ matter when network training converges, rather than just at the initialization. For the orthogonal case, our empirical results (cf. Figure \ref{FigIllus}) show that as the training proceeds, singular value spectra of weight matrices diverge from their initial conditions. We are thus motivated to investigate along this line, from the empirical observations of Figure \ref{FigIllus} and also the theoretical analysis in \cite{ExactSolution}.

More specifically, we propose a simple yet very effective network training method, which preserves the orthogonality of weight matrices during the procedure of network training. This amounts to solving the following constrained optimization problem
\begin{eqnarray}\label{EqnConstrainedSVB}
\min_{\Theta = \{ \mathbf{W}^l, \mathbf{b}^l\}_{l=1}^L} {\cal{L}}\left( \{\mathbf{x}_i, \mathbf{y}_i\}_{i=1}^K; \Theta\right) \nonumber \\ \mathrm{s.t.} \ \mathbf{W}^l \in {\cal{O}} \ \forall \ l\in \{1, \dots, L\} ,
\end{eqnarray}
where $\cal{O}$ stands for the set of matrices whose row or column vectors are orthogonal (or near orthogonal). Compared with standard SGDs, the feasible set of problem (\ref{EqnConstrainedSVB}) for $\{ \mathbf{W}^l\}_{l=1}^L$ is much reduced. We approximately solve this problem based on SGD (or its variants): we simply bound, after every $T_{svb}$ iterations of SGD training, all the singular values of each $\mathbf{W}^l$, for $l = 1, \dots, L$, in a narrow band $[1/(1+\epsilon), (1+\epsilon)]$ around the value of $1$, where $\epsilon$ is a specified small constant. Algorithm \ref{AlgmSVB} presents details of our proposed Singular Value Bounding (SVB) method. In Section \ref{SecTechAnalysis}, we present theoretical analysis on deep linear networks to justify the advantages of our proposed SVB on \emph{forward propagation to achieve training objectives}, and \emph{backward propagation of training errors}. In Section \ref{SecBNCompabibility}, we prove that BN could cause an ill-conditioned layer transform. To improve BN and make BN be compatible with SVB, we propose Bounded Batch Normalization (BBN) (cf. Algorithm \ref{AlgmBBN}), which removes such a risk by directly controlling the learning of BN parameters. Experiments in Section \ref{SecExps} on image classification show that SVB improves over SGD based methods, and in many cases with a large margin. And BBN further improves the performance on deep networks with BN layers.


\noindent\textbf{Empirical computation cost} Applying SVB to network training amounts to solving singular value decompositions (SVD) for weight matrices of all the network layers. We note that this cost can be amortized by doing SVB every $T_{svb}$ number of iterations. We usually apply SVB once every epoch of SGD training (for CIFAR10 with a batch size of $128$, this amounts to doing SVB once every $391$ iterations). The wall-clock time caused by SVB is practically negligible. In fact, we often observe even faster training when using SVB, possibly due to the better conditioning of weight matrices resulting from SVB.

\begin{algorithm}[t]
{\footnotesize

\SetKwInOut{Input}{input}
\SetKwInOut{Output}{output}

\Input{A network of $L$ layers with trainable parameters $\Theta = \{ \mathbf{W}^l, \mathbf{b}^l \}_{l=1}^L$, training loss $\cal{L}$, learning rate $\eta$, the maximal number $T$ of training iterations, a specified number $T_{svb}$ of iteration steps, a small constant $\epsilon$}

Initialize $\Theta$ such that $\mathbf{W}^{l\top}\mathbf{W}^l = \mathbf{I}$ or $\mathbf{W}^{l}\mathbf{W}^{l\top} = \mathbf{I}$ for $l = 1, \dots, L$

\For{$t = 0, \dots, T-1$}{
Update $\Theta_{t+1} \leftarrow \Theta_t - \eta \frac{\partial{\cal{L}}}{\partial{\Theta_t}}$ using SGD based methods

\While{training proceeds for every $T_{svb}$ iterations}{

\For{$l=1, \dots, L$}{
Perform $[\mathbf{U}^l, \mathbf{S}^l, \mathbf{V}^l] = \mathrm{svd}(\mathbf{W}^l)$

Let $\{ s_i^l \}_{i=1}^{N_l}$ be the diagonal entries of $\mathbf{S}^l$

\For{$i = 1, \dots, N_l$}{
$s_i^l = 1+\epsilon \ \ \text{if} \ \ s_i^l > 1 + \epsilon$

$s_i^l = 1/(1+\epsilon) \ \ \text{if} \ \ s_i^l < 1/(1+\epsilon)$
}
}

\If{network contains BN layers}{Use BBN of Algorithm \ref{AlgmBBN} to update BN parameters}

}
}

\Output{Trained network with parameters $\Theta_T$ for inference}

\caption{Singular Value Bounding} \label{AlgmSVB}

}
\end{algorithm}

\section{Propagations of all directions of variations with Singular Value Bounding}
\label{SecTechAnalysis}

In this section, we present theoretical analysis on deep linear networks to discuss the importance of forward-propagating \emph{all the directions} of training objectives and backward-propagating those of training errors, in order to better train deep neural networks. Our analyses resemble, but are different from, those in \cite{ExactSolution} (cf. Section \ref{SecLitecture} for details of the difference). These analyses justify our proposed SVB algorithm, and are supported by the experimental results reported in Section \ref{SecExps}.



\subsection{The forward propagation}
\label{SecDeepLinearTechAnalysisForward}


We start our analysis of optimal network solutions with a simple two-layer linear network that computes $\mathbf{W}^2\mathbf{W}^1\mathbf{x}$, where we have used a linear activation $f(\mathbf{z}) = \mathbf{z}$ and ignored the bias for simplicity. Using squared Euclidean distance as the training criterion gives the following loss function ${\cal{L}} = \frac{1}{2K} \sum_{i=1}^K \| \mathbf{y}_i - \mathbf{W}^2\mathbf{W}^1\mathbf{x}_i \|_2^2 $. To minimize $\cal{L}$ with respect to (w.r.t.) $\mathbf{W}^1$ and $\mathbf{W}^2$, we note that the optimal solutions are characterized by the gradients
\begin{eqnarray}\label{EqnGradW1W2}
\frac{\partial{\cal{L}}}{ \partial{\mathbf{W}^1}} = \mathbf{W}^{2\top} \left( \mathbf{C}^{yx} - \mathbf{W}^2\mathbf{W}^1\mathbf{C}^{xx} \right) \nonumber \\
\frac{\partial{\cal{L}}}{ \partial{\mathbf{W}^2}} = \left( \mathbf{C}^{yx} - \mathbf{W}^2\mathbf{W}^1\mathbf{C}^{xx} \right) \mathbf{W}^{1\top}  ,
\end{eqnarray}
where $\mathbf{C}^{yx} = \frac{1}{K}\sum_{i=1}^K \mathbf{y}_i\mathbf{x}_i^{\top}$ and $\mathbf{C}^{xx} = \frac{1}{K}\sum_{i=1}^K \mathbf{x}_i\mathbf{x}_i^{\top}$. When training deep networks, the input samples $\{\mathbf{x}_i\}_{i=1}^K$ are usually pre-processed by whitening, i.e., each $\mathbf{x}_i$ has zero mean and $\mathbf{C}^{xx} = \mathbf{I}$. With input data whitening, $\mathbf{C}^{yx}$ is in fact the cross-covariance matrix between input and output training samples, which models how the input variations relate to those of the outputs. Thus $\mathbf{C}^{yx}$ contains all the information that determines the learning results of (\ref{EqnGradW1W2}) w.r.t. $\mathbf{W}_1$ and $\mathbf{W}_2$. Applying SVD to $\mathbf{C}^{yx}$ gives $\mathbf{C}^{yx} = \mathbf{U}^y \mathbf{S}^{yx} \mathbf{V}^{x\top}$, where the orthogonal matrix $\mathbf{U}^y \in \mathbb{R}^{N_y\times N_y}$ contains columns of singular vectors in the output space that represent \emph{independent directions of output variations}, the orthogonal matrix $\mathbf{V}^{x} \in \mathbb{R}^{N_x\times N_x}$ contains columns of singular vectors in the input space that represent \emph{independent directions of input variations}, and $\mathbf{S}^{yx} \in \mathbb{R}^{N_y\times N_x}$ is a diagonal matrix with ordered singular values $ \sigma_1 \ge \sigma_2 \ge \cdots \ge \sigma_{\min(N_x, N_y)}$.

As suggested in \cite{ExactSolution}, when we initialize $\mathbf{W}^1$ and $\mathbf{W}^2$ as
\begin{eqnarray}\label{EqnW1W2InitCondition}
\mathbf{W}^1 = \mathbf{R}\mathbf{S}^1\mathbf{V}^{x\top}, \ \ \mathbf{W}^2 = \mathbf{U}^y\mathbf{S}^2\mathbf{R}^{\top},
\end{eqnarray}
where $\mathbf{R} \in \mathbb{R}^{N_1\times N_1}$ is an arbitrary orthogonal matrix and $\mathbf{S}^1$ and $\mathbf{S}^2$ are diagonal matrices with nonnegative entries, and keep $\mathbf{R}$ fixed during optimization, the gradients (\ref{EqnGradW1W2}) at optimal solutions can be derived as
\begin{eqnarray}\label{EqnGradW12ToSigma12}
\frac{\partial{\cal{L}}}{ \partial{\mathbf{W}^1}} = \mathbf{R}\mathbf{S}^{2\top} \left( \mathbf{S}^{yx} - \mathbf{S}^2\mathbf{S}^1 \right) \mathbf{V}^{x\top} \nonumber \\
\frac{\partial{\cal{L}}}{ \partial{\mathbf{W}^2}} = \mathbf{U}^y \left( \mathbf{S}^{yx} - \mathbf{S}^2\mathbf{S}^1 \right) \mathbf{S}^{1\top} \mathbf{R}^{\top} .
\end{eqnarray}
Since $\mathbf{R}$ is fixed, the above conditions ensure that $\mathbf{W}^1$ and $\mathbf{W}^2$ are optimized along their respective independent directions of variations. Denote $s_m$ and $t_m$, $m = 1, \dots, \min(N_y, N_1, N_x)$, are the $m^{th}$ diagonal entries of $\mathbf{S}^1$ and $\mathbf{S}^2$ respectively. By change of optimization variables, (\ref{EqnGradW1W2}) can be further simplified as the following equations for each $m^{th}$ direction of variations
\begin{eqnarray}\label{EqnGradSigma1Sigma2}
\frac{\partial{\cal{L}}}{\partial{s_m}} = \left( \sigma_m - s_m t_m \right) t_m , \ \ \frac{\partial{\cal{L}}}{\partial{t_m}} = \left( \sigma_m - s_m t_m \right) s_m .
\end{eqnarray}
In fact, the gradients (\ref{EqnGradSigma1Sigma2}) w.r.t. $s_m$ and $t_m$ arise from the following energy function
\begin{eqnarray}\label{EqnTwoLayerEnergyFunc}
{\cal{E}}(s_m, t_m) = \frac{1}{2} \left( \sigma_m - s_m t_m \right)^2 ,
\end{eqnarray}
showing that the product of optimal pairs $s_m$ and $t_m$ approaches $\sigma_m$.

We subsequently extend the analysis from (\ref{EqnGradW1W2}) to (\ref{EqnTwoLayerEnergyFunc}) for a deep linear network of $L$ layers. With the same loss function $\cal{L}$ of squared Euclidean distance, the optimal weight matrix $\mathbf{W}^l$ of the $l^{th}$ layer is characterized by the gradient
\begin{eqnarray}\label{EqnGradWl}
\frac{\partial{\cal{L}}}{ \partial{\mathbf{W}^l}} = \left( \prod_{i=l+1}^L \mathbf{W}^i \right)^{\top} \left( \mathbf{C}^{yx} - \prod_{i=1}^L \mathbf{W}^i \right) \left(\prod_{i=1}^{l-1} \mathbf{W}^i\right)^{\top} ,
\end{eqnarray}
where $\prod_{i=l}^{l'}\mathbf{W}^i = \mathbf{W}^{l'}\mathbf{W}^{l'-1}\cdots \mathbf{W}^l$ with the special case that $\prod_{i=l}^{l'}\mathbf{W}^i = \mathbf{I}$ when $l > l'$, and we have assumed in (\ref{EqnGradWl}) that $\mathbf{C}^{xx} = \mathbf{I}$. Similar to (\ref{EqnW1W2InitCondition}), when we initialize weight matrices of the deep network as $\mathbf{W}^l = \mathbf{R}^{l+1}\mathbf{S}^l\mathbf{R}^{l\top}$ for any $l \in \{1, \dots, L\}$, where each $\mathbf{R}^l$ is an orthogonal matrix with the special cases that $\mathbf{R}^1 = \mathbf{V}^x$ and $\mathbf{R}^{L+1} = \mathbf{U}^y$, and each $\mathbf{S}^l$ is an diagonal matrix with nonnegative entries, and keep $\{ \mathbf{R}^l \}_{l=1}^{L+1}$ fixed during optimization \footnote{Alternatively, one might relax this constraint and update $\{\mathbf{W}^l\}_{l=1}^L$ using standard methods such as SGD, and change the left and right singular vectors of each updated $\mathbf{W}^l$ to satisfy $\mathbf{W}^l = \mathbf{R}^{l+1}\mathbf{S}^l\mathbf{R}^{l\top}$ (with varying sets of $\{\mathbf{R}^l\}_{l=1}^{L+1}$). However, this would cause mixing of different directions in the connecting output/input spaces across layers.}, the gradient (\ref{EqnGradWl}) at optimal solutions can be derived as
\begin{equation}\label{EqnGradWlToSigmal}
\begin{aligned}
\frac{\partial{\cal{L}}}{ \partial{\mathbf{W}^l}} = \mathbf{R}^{l+1} \left( \prod_{i=l+1}^L \mathbf{S}^i \right)^{\top} \left( \mathbf{S}^{yx} - \prod_{i=1}^L \mathbf{S}^i \right) \left(\prod_{i=1}^{l-1} \mathbf{S}^i\right)^{\top} \mathbf{R}^{l\top} .
\end{aligned}
\end{equation}
By change of optimization variables, (\ref{EqnGradWlToSigmal}) can be further simplified as the following independent gradient for the $m^{th}$ direction of variations with $m \le M = \min(N_y, \dots, N_l, \dots, N_x)$
\begin{eqnarray}\label{EqnGradSigmal}
\frac{\partial{\cal{L}}}{ \partial{s_m^l}} = \prod_{i=l+1}^L s_m^i \left( \sigma_m - \prod_{i=1}^L s_m^i \right) \prod_{i=1}^{l-1} s_m^i,
\end{eqnarray}
which turns out to be the gradient of the energy function
\begin{eqnarray}\label{EqnDeepLinearNetEnergyFunc}
{\cal{E}}(s_m^1,\dots,s_m^L) = \frac{1}{2} \left( \sigma_m - \prod_{l=1}^L s_m^l \right)^2 .
\end{eqnarray}

The positive scalar $\sigma_m$ in (\ref{EqnDeepLinearNetEnergyFunc}) represents \emph{the strength of the $m^{th}$ direction of input-output correlations}. It is usually fixed given provided training data. To characterize the conditions under which the minimum energy of (\ref{EqnDeepLinearNetEnergyFunc}) can be achieved, denote $s_m^{l_{\max}} = \max(s_m^1, \dots, s_m^L)$ and $s_m^{l_{\min}} = \min(s_m^1, \dots, s_m^L)$. One can easily prove that when $L \rightarrow \infty$, it is \emph{necessary} that $s_m^{l_{\max}} > 1$ and $s_m^{l_{\min}} < 1$. Conversely, the \emph{sufficient} conditions for \emph{not achieving} the minimum energy of (\ref{EqnDeepLinearNetEnergyFunc}) are either $s_m^{l_{\max}} < 1$ or $s_m^{l_{\min}} > 1$, when $L \rightarrow \infty$.

For any fixed and finite $\sigma_m$, our proposed SVB algorithm is potentially able to achieve the minimum energy of (\ref{EqnDeepLinearNetEnergyFunc}) (although it does not meet the assumptions used to derive (\ref{EqnDeepLinearNetEnergyFunc})), by choosing an appropriate value of $\epsilon$ so that values of $\{ s_m^l \}_{l=1}^L$ are properly learned to range in a narrow band $\left[ 1/(1+\epsilon), 1+\epsilon\right]$. This applies to any of the $M$ directions of input-output correlations. Existing network training methods have no such constraints, and $\{ s_m^l \}$ of all layers/directions are free to be scaled up or down, resulting in very uneven magnitude distribution of $\{ \{ s_m^l \}_{l=1}^L \}_{m=1}^M$. Consequently, training easily falls in local minima that minimize (\ref{EqnDeepLinearNetEnergyFunc}) for certain directions, but not for all of the $M$ ones. And only parts of the input-output correlations are taken into account during learning.

Our derivation from (\ref{EqnGradWl}) to (\ref{EqnGradWlToSigmal}) requires that the output singular vectors of the weight matrix of layer $l$ be the input singular vectors of that of layer $l+1$. However, it does not hold true in the SGD based Algorithm \ref{AlgmSVB}, where weight matrices are updated without such constraints. Consider a two-layer basic component $\mathbf{W}^{l+1} \mathbf{W}^l$ in (\ref{EqnGradWl}), which propagates signal activations (and hence information of input variations) from layer $l$ to layer $l+1$. After SGD updating, Algorithm \ref{AlgmSVB} computes SVDs of the updated $\mathbf{W}^{l+1}$ and $\mathbf{W}^l$, resulting in $\mathbf{W}^{l+1} \mathbf{W}^l = \mathbf{U}^{l+1} \mathbf{S}^{l+1} \mathbf{V}^{l+1\top} \mathbf{U}^l \mathbf{S}^l \mathbf{V}^{l\top}$. While one may initialize $\mathbf{W}^{l+1}$ and $\mathbf{W}^l$ such that $\mathbf{V}^{l+1} = \mathbf{U}^l$, after SGD updating, they are generally not equal. Denote $\mathbf{M} = \mathbf{S}^{l+1} \mathbf{V}^{l+1\top} \mathbf{U}^l \mathbf{S}^l$, we have
\begin{eqnarray}\label{EqnStrengthMixing}
\mathbf{M}_{m, m'} = s_m^{l+1}s_{m'}^l \mathbf{v}_m^{l+1\top} \mathbf{u}_{m'}^l ,
\end{eqnarray}
where $\mathbf{M}_{m, m'}$ is the $(m, m')$ entry of $\mathbf{M}$, $\mathbf{v}_m^{l+1}$ is the $m^{th}$ column of $\mathbf{V}^{l+1}$, $\mathbf{u}_{m'}^l$ is the $m'^{th}$ column of $\mathbf{U}^l$, and $s_m^{l+1}$ and $s_{m'}^l$ are respectively the $m^{th}$ and $m'^{th}$ singular values of $\mathbf{S}^{l+1}$ and $\mathbf{S}^l$. By projecting $\mathbf{u}_{m'}^l$ onto $\mathbf{v}_m^{l+1}$, $\mathbf{v}_m^{l+1\top} \mathbf{u}_{m'}^l$ represents the \emph{mixing} of the $m'^{th}$ direction of variations in the output space of layer $l$ with the $m^{th}$ one in the input space of layer $l+1$. By bounding $s_m^{l+1}$ and $s_{m'}^l$, our proposed SVB algorithm controls both the \emph{independent} (when $m = m'$ and the assumptions from (\ref{EqnGradWl}) to (\ref{EqnGradWlToSigmal}) hold), and the \emph{mixing} strengths of propagation across layers. Without such constraints, some directions of variations could be over-amplified while others are strongly attenuated, when signals are propagated from lower layers to higher layers.


\subsection{The backward propagation}
\label{SecDeepLinearTechAnalysisBackward}

For a deep linear network that performs cascaded computations of $ \mathbf{x}^l = \mathbf{W}^l\mathbf{x}^{l-1}$ for $l=1, \dots, L$, the gradient of loss function $\cal{L}$ w.r.t. the output activation $\mathbf{x}^l$ of layer $l$ is written as
\begin{eqnarray}\label{EqnGradWRTXl}
\frac{\partial{\cal{L}}}{ \partial{\mathbf{x}^l}} = \left( \frac{\partial{\mathbf{x}^L}}{\partial{\mathbf{x}^l}} \right)^{\top} \frac{\partial{\cal{L}}}{\partial{\mathbf{x}^L}} = \left( \prod_{i=l+1}^L \mathbf{W}^i \right)^{\top} \frac{\partial{\cal{L}}}{\partial{\mathbf{x}^L}} ,
\end{eqnarray}
where $\frac{\partial{\cal{L}}}{\partial{\mathbf{x}^L}}$ contains the error vector for back-propagation. For any $i \in \{l+1, \dots, L\}$, assume $\mathbf{W}^i$ satisfies the condition $\mathbf{W}^i = \mathbf{R}^{i+1}\mathbf{S}^i\mathbf{R}^{i\top}$ that we have used to derive from (\ref{EqnGradWl}) to (\ref{EqnDeepLinearNetEnergyFunc}), with analysis similar to the forward propagation case, we have
\begin{eqnarray}\label{EqnGradWRTXlSimp}
\frac{\partial{\cal{L}}}{ \partial{\mathbf{x}^l}} = \left( \sum_{m=1}^M \left(\prod_{i=l+1}^L s_m^i \right) \mathbf{r}_m^{L+1} \mathbf{r}_m^{l+1\top} \right)^{\top} \frac{\partial{\cal{L}}}{\partial{\mathbf{x}^L}} ,
\end{eqnarray}
where $\mathbf{r}_m^{L+1}$ (or $\mathbf{r}_m^{l+1}$) denotes the $m^{th}$ column of $\mathbf{R}^{L+1}$ (or $\mathbf{R}^{l+1}$), and $M = \min(N_L, \dots, N_l)$.
As the network goes deep (i.e., $L$ becomes large), $\prod_{i=l+1}^L s_m^i$ would either explode or vanish if $\{ s_m^i \}_{i=l+1}^{L}$ do not satisfy a necessary condition similar to the one for achieving the minimum of (\ref{EqnDeepLinearNetEnergyFunc}). Consequently, the $m^{th}$ \emph{component} error vector $\left(\prod_{i=l+1}^L s_m^i \right) \mathbf{r}_m^{l+1} \mathbf{r}_m^{L+1\top} \frac{\partial{\cal{L}}}{\partial{\mathbf{x}^L}} $ in (\ref{EqnGradWRTXlSimp}) would either explode or vanish. When $\epsilon \rightarrow 0$ in Algorithm \ref{AlgmSVB}, our proposed method guarantees that all the $M$ components of the error vector would propagate to lower layers without attenuation or explosion. In the ideal case of $N_L = \dots = N_l$, our method also guarantees that $\| \frac{\partial{\cal{L}}}{ \partial{\mathbf{x}^l}} \|_2 = \| \frac{\partial{\cal{L}}}{\partial{\mathbf{x}^L}} \|_2$, i.e., to preserve the norm of error vector. Without such constraints on singular values of $\{ \mathbf{W}^i \}_{i=l+1}^L$, it is still possible that the norm of error vector is preserved by amplifying some singular values while shrinking others, as the way advocated in \cite{XavierInit}. However, its norm preservation is achieved in a rather anisotropic way.

\section{Compatibility with Batch Normalization}
\label{SecBNCompabibility}

In this section, we investigate how our proposed network training algorithm could be compatible with Batch Normalization \cite{BatchNorm}. BN addresses a network training issue called \emph{internal covariate shift}, which slows down the training since distributions of each layer's inputs keep changing during the training process. BN alleviates this issue by inserting into network trainable normalization layers, which normalize each layer's neuron activations as zero mean and unit variance in a mini-batch and neuron-wise manner.

Formally, for a network layer computing $f(\mathbf{z}) = f(\mathbf{W}\mathbf{x}) \in \mathbb{R}^N$, BN inserts a normalization layer before the activation function, giving the new layer $f(\textrm{BN}(\mathbf{z})) = f(\textrm{BN}(\mathbf{W}\mathbf{x}))$, where we have ignored the bias term for simplicity. BN in fact applies the following linear transformation to $\mathbf{z}$
\begin{eqnarray}\label{EqnBNLinearTransform}
\textrm{BN}(\mathbf{z}) = \Gamma \Sigma (\mathbf{z} - \mathbf{\mu}) + \mathbf{\beta} ,
\end{eqnarray}
where each entry of $\mathbf{\mu} \in \mathbb{R}^{N}$ is the output mean at each of the $N$ neurons of the layer, the diagonal matrix $\Sigma \in \mathbb{R}^{N\times N}$ contains entries $\{1/\varsigma_i\}_{i=1}^N$ that is the inverse of the neuron-wise output standard deviation $\varsigma_i$ (obtained by adding a small constant to the variance for numerical stability), $\Gamma \in \mathbb{R}^{N\times N}$ is a diagonal matrix containing trainable scalar parameters $\{ \gamma_i \}_{i=1}^N$, and $\mathbf{\beta} \in \mathbb{R}^N$ is a trainable bias term. Note that during training, $\mu$ and $\varsigma$ for each neuron are computed using mini-batch samples, and during inference they are fixed representing the statistics of all the training population, which are usually obtained by running average. Thus the computation (\ref{EqnBNLinearTransform}) for each sample is deterministic after network training.

Inserting $\mathbf{z} = \mathbf{W}\mathbf{x}$ into (\ref{EqnBNLinearTransform}) we get
\begin{eqnarray}\label{EqnBNLinearTransformEquivalent}
\textrm{BN}(\mathbf{x}) = \widetilde{\mathbf{W}}\mathbf{x} + \tilde{\mathbf{b}} \ \ \textrm{s.t.} \ \ \widetilde{\mathbf{W}} = \Gamma\Sigma\mathbf{W} \ \ \tilde{\mathbf{b}} = \mathbf{\beta} - \Gamma\Sigma\mathbf{\mu} ,
\end{eqnarray}
which is simply a standard network layer with change of variables. The following lemma suggests that we may bound the entries $\{ \gamma_i/\varsigma_i \}_{i=1}^N$ of the product of the diagonal matrices $\Gamma$ and $\Sigma$, to make our proposed SVB algorithm be compatible with BN.

\vspace{0.2cm}
\noindent\textbf{Lemma 1} \emph{For a matrix $\mathbf{W} \in \mathbb{R}^{M\times N}$ with singular values of all $1$, and a diagonal matrix $\mathbf{G} \in \mathbb{R}^{M\times M}$ with nonzero entries $\{ g_i \}_{i=1}^M$, let $g_{\max} = \max(|g_1|, \dots, |g_M|)$ and $g_{\min} = \min(|g_1|, \dots, |g_M|)$, the singular values of $\widetilde{\mathbf{W}} = \mathbf{G}\mathbf{W}$ is bounded in $[g_{min}, g_{\max}]$. When $\mathbf{W}$ is fat, i.e., $M \le N$, and $\textrm{rank}(\mathbf{W}) = M$, singular values of $\widetilde{\mathbf{W}}$ are exactly $\{ |g_i| \}_{i=1}^M$.}
\vspace{0.2cm}

Proof of the lemma is given in Appendix \ref{ApendixSecBBNProof}. Lemma 1 suggests that for a deep network with BN layers, the trainable parameters $\{ \gamma_i \}_{i=1}^N$, together with sample statistics $\{\varsigma_i\}_{i=1}^N$, could change the conditioning of layer transform, and consequently the behaviors of signal propagation across network layers. In particular, when absolute values $\{ |\gamma_i/\varsigma_i| \}_{i=1}^N$ of the diagonal entries of $\Gamma\Sigma$ \emph{for all the network layers simultaneously drift up or down} away from the value of $1$, signal propagation would be susceptible to explosion or attenuation when the network goes deep. One direct way to remove this risk is to control the values of $\{ \gamma_i/\varsigma_i \}_{i=1}^N$, e.g., to let them be around $1$. However, this would also remove an important benefit of BN. More specifically, the introduction of trainable scaling parameters $\{ \gamma_i \}_{i=1}^N$ in BN is to make sure that after neuron-wise normalization by $\{\varsigma_i\}_{i=1}^N$ (and $\{\mu_i\}_{i=1}^N$), the change to layer outputs is compensated by $\{ \gamma_i \}_{i=1}^N$, so that the BN transform is overall an \emph{identity transform} \cite{BatchNorm}. One might expect that the value of each $\varsigma_i$ in $\Sigma$ is similar to that of the corresponding $\gamma_i$ in $\Gamma$. However, this is not the case in practice. In fact, $\{ \gamma_i \}_{i=1}^N$ bring additional and significant benefits to training of deep neural networks: the decoupled $\{ \gamma_i \}_{i=1}^N$ enable scales of the magnitude of features at different network layers become freely adjustable for better training objectives. This advantage is also leveraged in recent works such as \cite{WeightNorm} to improve network training. Inspired by this scheme of BN, we introduce a decoupled scalar $\alpha$ from $\Gamma\Sigma$, and propose to control the re-scaled version $\{ \frac{1}{\alpha} \gamma_i/\varsigma_i \}_{i=1}^N$, instead of $\{ \gamma_i/\varsigma_i \}_{i=1}^N$, to make our proposed SVB be compatible with BN. We set $\alpha = \frac{1}{N}\sum_{i=1}^N \gamma_i/\varsigma_i$ during network training. Note that re-scaling the magnitude scales of features at different layers is equivalent to simultaneously scaling up $\{ s_m^l\}_{m=1}^{M}$ of all the $M$ directions in (\ref{EqnDeepLinearNetEnergyFunc}) for certain layers, while simultaneously scaling down for other layers, and this does not cause sacrifice of propagation of certain directions of input-output correlations. Algorithm \ref{AlgmBBN} presents our improved BN transform called \emph{Bounded Batch Normalization (BBN)}. We note that in Algorithm \ref{AlgmBBN}, we do not take the absolute values. This is because values of $\{ \gamma_i \}_{i=1}^N$ are usually initialized as $1$, and they are empirically observed to keep positive during the process of network training. Experiments in Section \ref{SecExps} show that image classification results are improved when using BBN instead of BN, demonstrating a consistency between our theoretical analysis and practical results.

\begin{algorithm}[t]
{\footnotesize

\SetKwInOut{Input}{input}
\SetKwInOut{Output}{output}

\Input{A network with $L$ BN layers, trainable parameters $\{ \Gamma_t^l \}_{l=1}^L$, $\{ \mathbf{\beta}_t^l \}_{l=1}^L$, and statistics $\{ \mathbf{\mu}_t^l \}_{l=1}^L$, $\{ \Sigma_t^l \}_{l=1}^L$ of BN layers at iteration $t$, a small constant $\tilde{\epsilon}$ }

Update to get $\{ \Gamma_{t+1}^l \}_{l=1}^L$ from $\{ \Gamma_t^l \}_{l=1}^L$ (and $\{ \mathbf{\beta}_{t+1}^l \}_{l=1}^L$ from $\{ \mathbf{\beta}_t^l \}_{l=1}^L$), using SGD based methods

Update to get $\{ \Sigma_{t+1}^l \}_{l=1}^L$ from $\{ \Sigma_t^l \}_{l=1}^L$ (and $\{ \mathbf{\mu}_{t+1}^l \}_{l=1}^L$ from $\{ \mathbf{\mu}_t^l \}_{l=1}^L$), using running average over statistics of mini-batch samples

\For{$l = 1, \dots, L$}{
Let $\{ \gamma_i \}_{i=1}^{N_l}$ and $\{1/\varsigma_i\}_{i=1}^{N_l}$ be respectively the diagonal entries of $\Gamma_{t+1}^l$ and $\Sigma_{t+1}^l$

Let $\alpha = \frac{1}{N_l}\sum_{i=1}^{N_l} \gamma_i/\varsigma_i$

\For{$i = 1, \dots, N_l$}{

$\gamma_i = \alpha\varsigma_i(1+\tilde{\epsilon}) \ \ \text{if} \ \ \frac{1}{\alpha}\gamma_i/\varsigma_i > 1 + \tilde{\epsilon}$

$\gamma_i = \alpha\varsigma_i/(1+\tilde{\epsilon)} \ \ \text{if} \ \ \frac{1}{\alpha}\gamma_i/\varsigma_i < 1/(1+\tilde{\epsilon)}$

}

}

\Output{Updated BN parameters and statistics at iteration $t+1$}

\caption{Bounded Batch Normalization} \label{AlgmBBN}

}
\end{algorithm}


\section{Experiments}
\label{SecExps}

In this section, we present image classification results to show the efficacy of our proposed SVB and BBN algorithms for training deep neural networks. We use benchmark datasets including CIFAR10, CIFAR100 \cite{Cifar}, and ImageNet \cite{ILSVRC15}. CIFAR10 is intensively used for our controlled studies. We investigate how SVB and BBN perform on standard ConvNets, and also the modern architectures of (pre-activation versions of) ResNets \cite{PreActResNet}, Wide ResNets \cite{WideResNet}, and Inception-ResNets \cite{InceptionResNet}.

We use BN layers or our proposed BBN layers in all networks. Training is based on SGD with momentum using softmax loss function. We initialize networks using orthogonal weight matrices (cf. Algorithm \ref{AlgmSVB}). Except experiments reported in Table \ref{TableWideResNetComp}, all other experiments are based on a mini-batch size of $128$, momentum of $0.9$, and weight decay of $0.0001$; the learning rate starts from $0.5$ and ends at $0.001$, and decays every two epochs until the end of $160$ epoches of training. When SVB is turned on, we apply it to weight matrices of all layers after every epoch of training. This amounts to performing SVB every $391$ iterations for CIFAR10 and CIFAR100. We will present the effects of the bounding parameters $\epsilon$ in Algorithm \ref{AlgmSVB} and $\tilde{\epsilon}$ in Algorithm \ref{AlgmBBN} shortly.

\subsection{Controlled studies using ConvNets}
\label{SecExpsConvNetStudies}

In this section, we use ConvNets to study the behaviors of our proposed SVB algorithm on deep network training. We choose modern convolutional architectures from \cite{VGGNet,ResNet}. The networks start with a conv layer of $16$ $3\times 3$ filters, and then sequentially stack three types of $2X$ conv layers of $3\times 3$ filters, each of which has the feature map sizes of $32$, $16$, and $8$, and filter numbers $16$,  $32$,  and $64$, respectively. Spatial sub-sampling of feature maps is achieved by conv layers of stride $2$. The networks end with a global average pooling and fully-connected layers. Thus for each network we have $6X+2$ weight layers in total. We use networks of $X = 3$ and $X = 6$ for our studies, which give $20$- and $38$-layer networks respectively.

The used CIFAR10 dataset consists of $10$ object categories of $60,000$ color images of size $32\times 32$ ($50,000$ training and $10,000$ testing ones). We use raw images without pre-processing. The data augmentation follows the standard manner in \cite{DeeplySupervisedNet}: during training, we zero-pad $4$ pixels along each image side, and sample a $32\times 32$ region crop from the padded image or its horizontal flip; during testing, we simply use the original non-padded image.

\begin{figure*}[t]
\centering

\includegraphics[scale=0.4]{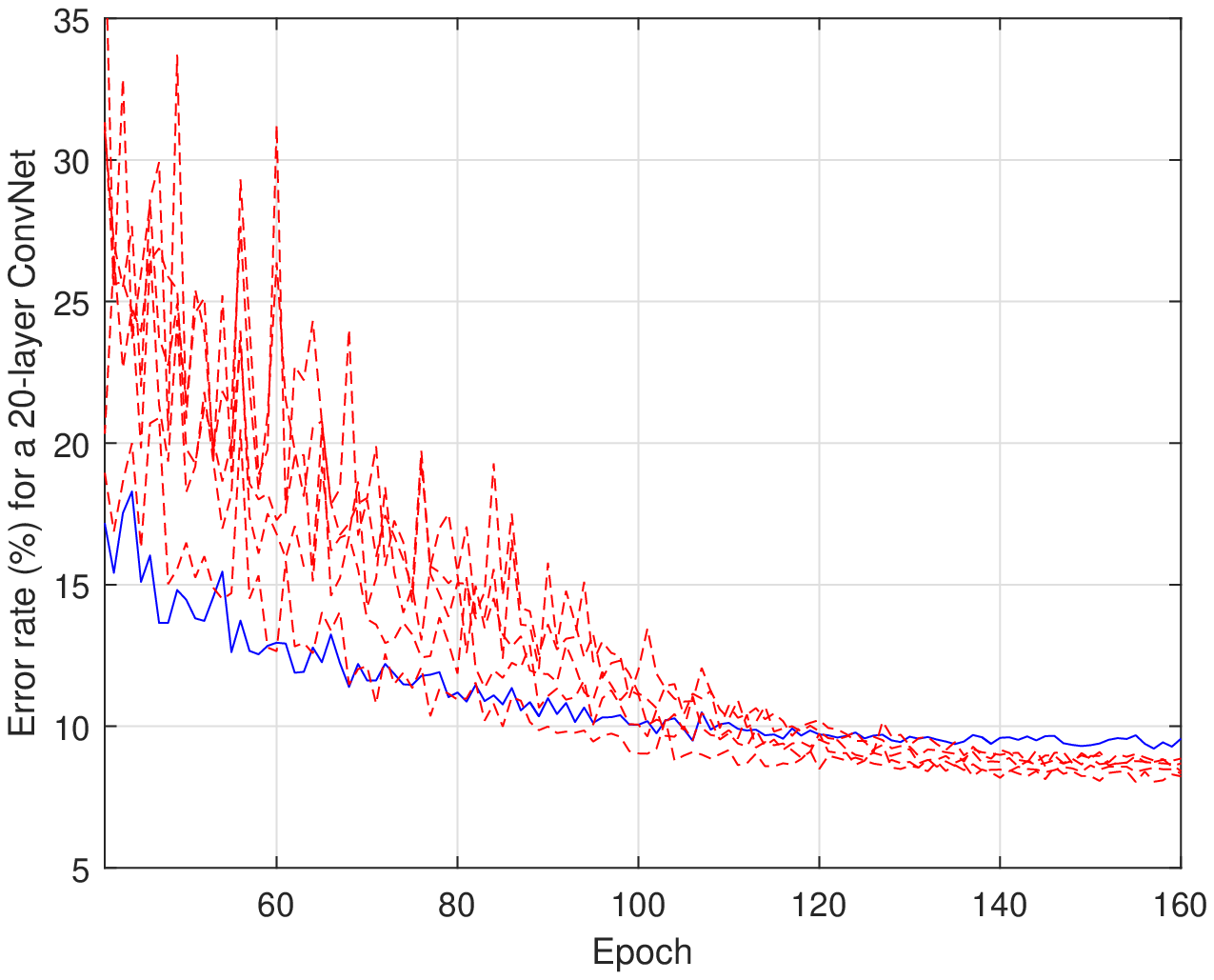}\includegraphics[scale=0.4]{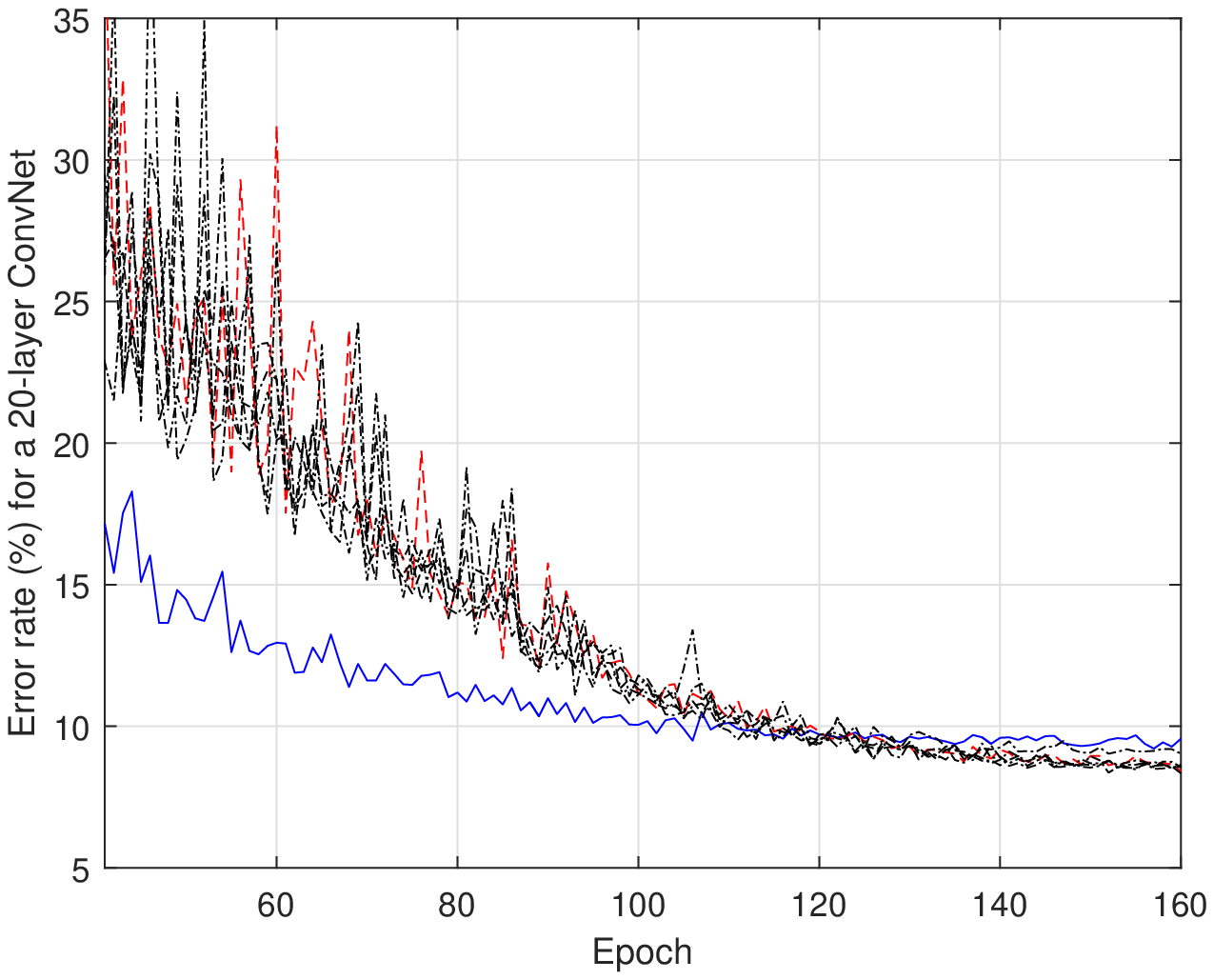}\includegraphics[scale=0.4]{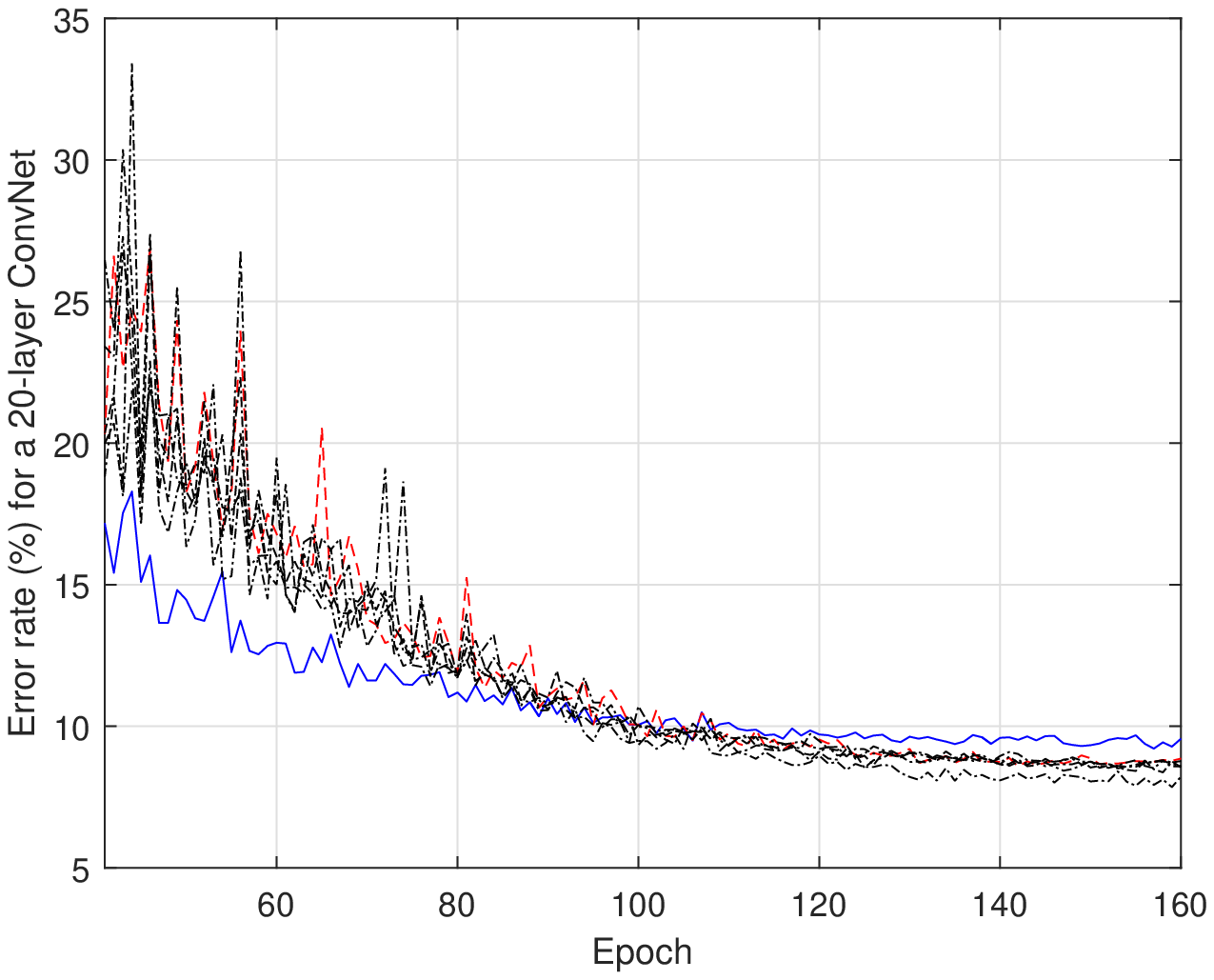} \\ \includegraphics[scale=0.4]{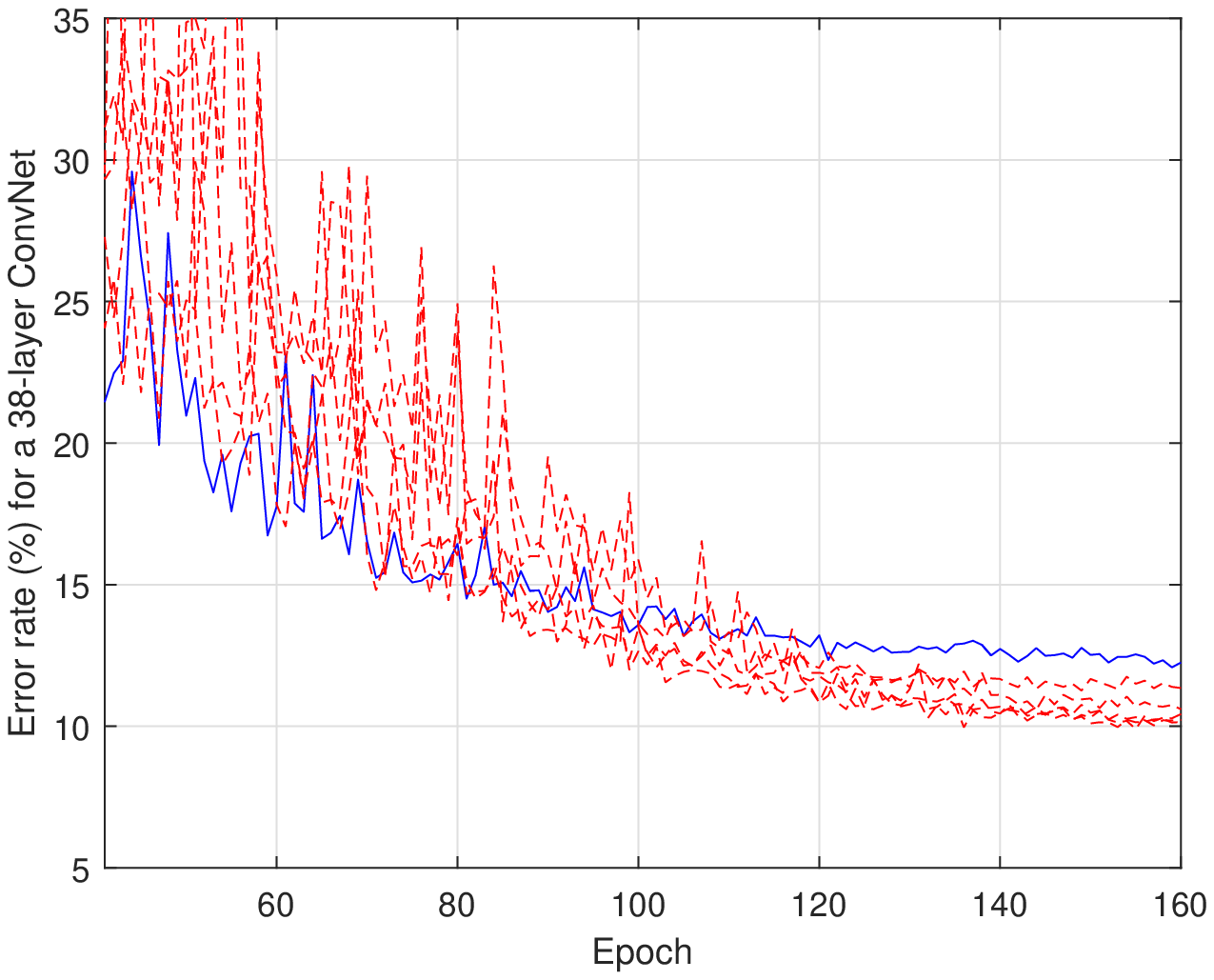}\includegraphics[scale=0.4]{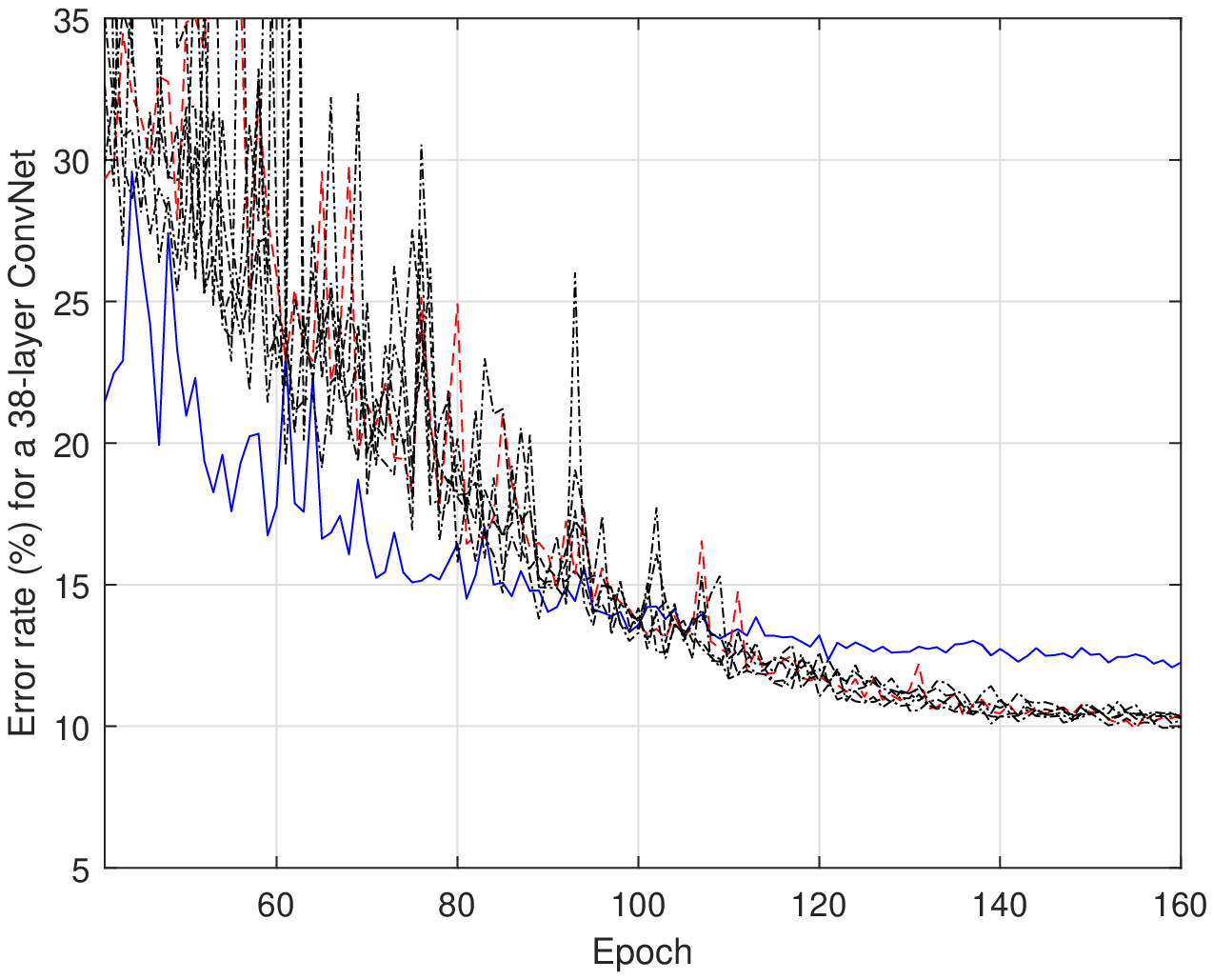}\includegraphics[scale=0.4]{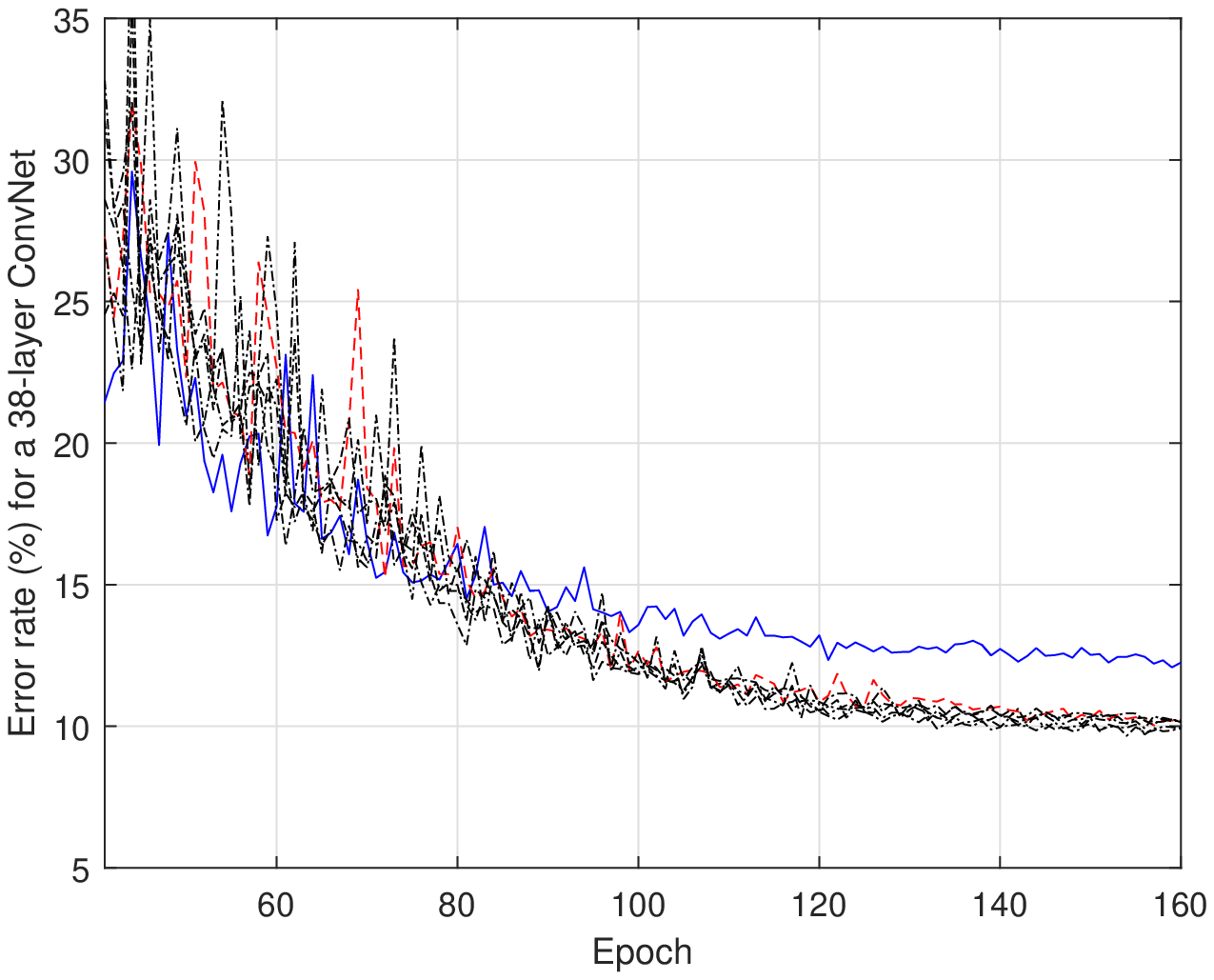} \\

\caption{ {\small Validation curves on CIFAR10 using two ConvNets of $20$ (first row) and $38$ (second row) weight layers respectively. Blue lines are results by SGD with momentum. Red lines are results by SVB at different values of $\epsilon$ ($0.01, 0.05, 0.2, 0.5, 1$) in Algorithm \ref{AlgmSVB}. Black lines are results using both SVB and BBN at different values of $\tilde{\epsilon}$ ($0.2, 0.5, 1, 2, 4$)) in Algorithm \ref{AlgmBBN}, where black line results in the middle and right columns are obtained by fixing $\epsilon$ of Algorithm \ref{AlgmSVB} as $0.05$ and $0.5$ respectively. These parameter settings are simply casual choices.  }  }  \label{FigExpControlStudies}
\end{figure*}

Figure \ref{FigExpControlStudies} shows that for each depth case, our results using SVB are consistently better than those from standard SGD with momentum. For $20$-layer network, SGD with momentum gives an error rate of $9.21$, and our best result using SVB improves the error rate to $8.03$. For $38$-layer network, SGD with momentum gives an error rate of $12.08$, and our best result using SVB improves the error rate to $9.90$. These results verify that bounding the singular values of weight matrices indeed improves the conditioning of layer transform.

Replacing BN with BBN further improves the result to $7.85$ for the $20$-layer network, and to $9.66$ for the $38$-layer network. The improvement is however less significant as compared with that from SGD with momentum to SVB. This shows that in the case of plain ConvNets, SVB has largely improved the conditioning of layer transform, and the ill-conditioning caused by BN is not severe. Indeed, in the more complex ResNet type architectures, BBN improves over BN effectively at a higher accuracy level, as presented shortly.

Comparative results in Figure \ref{FigExpControlStudies} also suggest that with the increase of network layers, training becomes more difficult: results of deeper network ($X = 6$) are worse than those of shallower one ($X = 3$). This is consistent with the observations in \cite{ResNet}. Although our SVB and BBN improve the performance, they do not solve this training difficulty of plain ConvNets.

\subsection{Ablation studies using ResNet}

We conduct experiments to investigate whether our proposed SVB and BBN methods are effective for ``residual learning'' \cite{ResNet}. We use an architecture similar to those presented in Section 4.2 in \cite{ResNet}, but change it to the pre-activation version \cite{PreActResNet}. The network construction is based on the ConvNets presented in Section \ref{SecExpsConvNetStudies}, and we use an ``identify shortcut'' to connect every two conv layers of $3\times 3$ filters, and use a ``projection shortcut'' when sub-sampling of feature maps is needed. We use a network of $X = 11$ for experiments in this section, which gives $68$ weight layers.

We conduct ablation studies by switching SVB on or off, and switching BBN on or off. The parameters $\epsilon$ and $\tilde{\epsilon}$ are fixed as $0.5$ and $1$ respectively. All experiments are run for $5$ times and we report both the best, and the mean and standard deviation results (in parenthesis) in Table \ref{TableResNetAblation}. These results show that using SVB improves deep residual learning, and BBN further improves over standard BN, demonstrating the efficacy of our proposed methods for modern deep architectures.

\begin{table}[t]
\caption{ {\small Ablation studies on CIFAR10, using a pre-activation ResNet with $68$ weight layers of $3\times 3$ convolutional filters. We run each setting for $5$ times, using standard data augmentation \cite{DeeplySupervisedNet}. Results are in the format of best (mean + std).  } } \label{TableResNetAblation}
\begin{center}
\begin{tabular}{cc}
\hline {\scriptsize Training methods} & {\scriptsize Error rate ($\%$)} \\
\hline {\scriptsize SGD with momentum + BN}   &  $6.10$ ($6.22 \pm 0.14$) \\
\hline {\scriptsize SVB + BN}                 &  $5.65$ ($5.79 \pm 0.10$) \\
\hline {\scriptsize SVB + BBN}                &  $\mathbf{5.37}$ ($5.49 \pm 0.11$)\\
\hline
\end{tabular}
\end{center}
\end{table}

\subsection{Comparisons with the state-of-the-art}
\label{SecExpInceptionResNetComp}

We apply our proposed methods to Wide ResNet \cite{WideResNet}, and a pre-activation version of Inception-ResNet \cite{InceptionResNet}, and compare with the state-of-the-art results on CIFAR10 and CIFAR100. The CIFAR100 dataset has the same number of $32\times 32$ color images as CIFAR10 does, but it has $100$ object categories and each category contains one tenth images of those of CIFAR10. We use raw data without pre-processing, and do data augmentation using the same manner as for CIFAR10.

Our architecture of Wide ResNet is the same as that of ``WRN-28-10'' in \cite{WideResNet}, but our training hyperparameters (cf. the beginning of Section \ref{SecExps}) are different from those in \cite{WideResNet}. Our architecture of Inception-ResNet is very similar to that reported in Figure 15 in \cite{InceptionResNet} (Inception-ResNet-v2), and we simply replace its ``stem'' module with three convolutional blocks (conv layers followed by BN and ReLU) that keep the $32\times 32$ image size but increase the number of feature maps from $3$ (the input color channels) to $384$, and also revise its other modules to the respective pre-activation versions according to \cite{PreActResNet}. When BBN is turned on, we replace all BN layers with the BBN ones. The parameters $\epsilon$ and $\tilde{\epsilon}$ are fixed as $0.5$ and $1$ respectively. Without using SVB and BBN, our architecture of Inception-ResNet gives an error rate of $5.40$ on CIFAR10 and $21.48$ on CIFAR100. With our proposed SVB and BBN, the results are significantly boosted to $4.17$ on CIFAR10 and $18.30$ on CIFAR100. Our proposed SVB and BBN further improve the result of Wide ResNet architecture on CIFAR10 to $3.58\%$.

To compare with the state-of-the-art method DenseNet \cite{DenseNet}, we use improved training hyperparameters inspired by \cite{DenseNet}: we use the batchsize of $64$ for CIFAR10 and $128$ for CIFAR100, and train for an extended duration of $300$ epochs; all other training parameters are the same as those described in the beginning of Section \ref{SecExps}. We set $\epsilon$ and $\tilde{\epsilon}$ of our methods as $0.5$ and $0.2$ respectively. Table \ref{TableWideResNetComp} reports the comparative results. Our results are based on the architectures proposed in \cite{WideResNet}: in Table \ref{TableWideResNetComp}, Wide ResNet uses the architecture ``WRN-28-10'' as in \cite{WideResNet}, and Wider ResNet increases the widening factor from 10 to 16 (corresponding to the architecture ``WRN-28-16'' in \cite{WideResNet}). Our proposed SVB and BBN improve both architectures, and achieve the new state-of-the-art results of $3.06$ on CIFAR10 and $16.90$ on CIFAR100. These results demonstrate the great potential of SVB and BBN on training modern deep architectures. \emph{In Table \ref{TableWideResNetComp}, Wider ResNet has much more model parameters than DenseNet does. However, we note that DenseNet practically consumes more memories in both training and inference. This is due to the architectural design of DenseNet: in (each stage/block of) DenseNet, the input of an upper layer is formed by concatenating output feature maps of all its lower layers; thus it is easy to have bottleneck layers of tremendous memory consumption when the number of each layer's output feature maps (i.e., the growth rate in \cite{DenseNet}) is large. This is indeed the case for model setting of the best result achieved by \cite{DenseNet}.}

\begin{table}[t]
\caption{ {\small Error rates ($\%$) of different methods on CIFAR10 and CIFAR100 \cite{Cifar}. All methods use standard data augmentation as in \cite{DeeplySupervisedNet}. A ``-'' indicates that result is not explicitly specified in the cited work. } } \label{TableInceptionResNetComp}
\begin{center}
\begin{tabular}{ccccc}
\hline {\scriptsize Methods}                              & {\scriptsize CIFAR$10$} & {\scriptsize CIFAR$100$} & {\scriptsize $\#$ layers} & {\scriptsize $\#$ params}  \\
\hline




{\scriptsize NIN \cite{NIN}}                              & $8.81$        &-              & - & - \\

{\scriptsize FitNet \cite{FitNet}}                        & $8.39$        &-              & $19$ & $2.5$M \\

{\scriptsize DSN \cite{DeeplySupervisedNet}}              & $7.97$        &-              & - & -  \\




{\scriptsize Highway \cite{HighwayNet}}                   & $7.54$        & $32.24$       & $19$ & $2.3$M \\

{\scriptsize ResNet \cite{ResNet}}                        & $6.43$        & $25.16$       & $110$ & $1.7$M  \\

{\scriptsize Stoc. Depth \cite{StochasticDepth}}          & $4.91$        & -             & $1202$ & $10.2$M \\

{\scriptsize Pre-Act ResNet \cite{PreActResNet}}          & $4.92$        & $22.71$       & $1001$ & $10.2$M  \\

{\scriptsize Wide ResNet \cite{WideResNet}}               & $4.17$  & $20.50$        & $28$ & $36.5$M \\

{\scriptsize ResNet of ResNet \cite{ResNetResNet}}        & $3.77$  & $19.73$        & - & $13.3$M \\

\hline
{\scriptsize Our Inception-ResNet}                            &          &           &       &    \\
{\scriptsize W/O SVB+BBN}                                     & $5.40$          & $21.48$          & $92$      &  $32.5$M \\
{\scriptsize Our Inception-ResNet}                           &   &   &     &    \\
{\scriptsize WITH SVB+BBN}                           & $4.17$ & $\mathbf{18.30}$ & $92$      &  $32.5$M \\

{\scriptsize Our Wide ResNet}                            &          &           &       &    \\
{\scriptsize W/O SVB+BBN}                                     & $4.50$          & $20.78$          & $28$      &  $36.5$M \\
{\scriptsize Our Wide ResNet}                           &   &   &     &    \\
{\scriptsize WITH SVB+BBN}                           & $\mathbf{3.58}$ & $18.32$ & $28$      &  $36.5$M \\
\hline
\end{tabular}
\end{center}
\end{table}

\begin{table}[t]
\caption{ {\small Comparisons of error rate ($\%$) with the state-of-the-art method DenseNet \cite{DenseNet} on CIFAR10 and CIFAR100 \cite{Cifar}. Our results are obtained by using improved training hyperparameters inspired by \cite{DenseNet}. All methods use standard data augmentation as in \cite{DeeplySupervisedNet}. \textbf{We note that DenseNet practically consumes more memories than Wider ResNet does.} } } \label{TableWideResNetComp}
\begin{center}
\begin{tabular}{ccccc}
\hline {\scriptsize Methods}                              & {\scriptsize CIFAR$10$} & {\scriptsize CIFAR$100$} & {\scriptsize $\#$ layers} & {\scriptsize $\#$ params}  \\
\hline

{\scriptsize DenseNet \cite{DenseNet}}                    & $3.46$  & $17.18$        & $190$ & $25.6$M \\

\hline

{\scriptsize Our Wide ResNet}                            &          &           &       &    \\
{\scriptsize W/O SVB+BBN}                                     & $3.78$          & $19.92$          & $28$      &  $36.5$M \\
{\scriptsize Our Wide ResNet}                           &   &   &     &    \\
{\scriptsize WITH SVB+BBN}                           & $\mathbf{3.24}$ & $17.47$ & $28$      &  $36.5$M \\

{\scriptsize Our Wider ResNet}                            &          &           &       &    \\
{\scriptsize W/O SVB+BBN}                                     & $3.64$          & $19.25$          & $28$      &  $94.2$M \\
{\scriptsize Our Wider ResNet}                           &   &   &     &    \\
{\scriptsize WITH SVB+BBN}                           & $\mathbf{3.06}$ & $\mathbf{16.90}$ & $28$      &  $94.2$M \\

\hline
\end{tabular}
\end{center}
\end{table}

\subsection{Preliminary results on ImageNet}

We present preliminary results on ImageNet \cite{ILSVRC15}, which has $1.28$ million images of $1000$ classes for training, and $50$ thousand images for validation. The data augmentation scheme follows \cite{InceptionResNet}. We investigate how SVB and BBN may help large-scaling learning, for which we use the pre-activation version of Inception-ResNet \cite{InceptionResNet}. We use the same parameter settings as those for the CIFAR10 experiments in Section \ref{SecExpInceptionResNetComp}, except the learning rate that starts from $0.045$. Table \ref{TableImageNet} shows that SVB and BBN indeed improve the large-scale learning, with a similar performance gain for top-1 and top-5 errors. The improvement is however lower than what we expected. We are interested for further studies in future research. We note that our architecture is almost identical to \cite{InceptionResNet}, but we did not manage to get the results in \cite{InceptionResNet}, possibly due to the different choices of gradient descent methods (\cite{InceptionResNet} uses RMSProp while ours are based on SGD with momentum).


\begin{table}[t]
\caption{ {\small Error rates of single-model and single-crop testing on the ImageNet validation set.  } } \label{TableImageNet}
\begin{center}
\begin{tabular}{ccc}
\hline {\scriptsize Training methods} & {\scriptsize Top-1 error ($\%$)} & {\scriptsize Top-5 error ($\%$)} \\
\hline {\scriptsize Our Inception-ResNet}   & $21.61$ & $5.91$  \\
\hline {\scriptsize Our Inception-ResNet WITH SVB+BN}   & $21.20$ & $5.57$  \\
\hline
\end{tabular}
\end{center}
\end{table}

\section{Conclusions}
\label{SecConclusion}

In this work, we present a simple yet effective method called Singular Value Bounding, to improve training of deep neural networks. SVB iteratively projects SGD based updates of network weights into a near orthogonal feasible set, by constraining all singular values of each weight matrix in a narrow band around the value of $1$. We further propose Bounded Batch Normalization, a method to remove the risk of ill-conditioned layer transform caused by batch normalization. We present theoretical analysis to justify our proposed methods. Experiments on benchmark image classification tasks show the efficacy.


\ifCLASSOPTIONcaptionsoff
  \newpage
\fi

\bibliographystyle{plain}
\bibliography{DeepLearning}

\appendices

\section{}
\label{ApendixSecBBNProof}

\noindent\textbf{Lemma 1} \emph{For a matrix $\mathbf{W} \in \mathbb{R}^{M\times N}$ with singular values of all $1$, and a diagonal matrix $\mathbf{G} \in \mathbb{R}^{M\times M}$ with nonzero entries $\{ g_i \}_{i=1}^M$, let $g_{\max} = \max(|g_1|, \dots, |g_M|)$ and $g_{\min} = \min(|g_1|, \dots, |g_M|)$, the singular values of $\widetilde{\mathbf{W}} = \mathbf{G}\mathbf{W}$ is bounded in $[g_{min}, g_{\max}]$. When $\mathbf{W}$ is fat, i.e., $M \le N$, and $\textrm{rank}(\mathbf{W}) = M$, singular values of $\widetilde{\mathbf{W}}$ are exactly $\{ |g_i| \}_{i=1}^M$.}

\begin{proof} We first consider the general case, and let $P = \min(M, N)$. Denote singular values of $\mathbf{W}$ as $\sigma_1 = \cdots = \sigma_P = 1$, and singular values of $\widetilde{\mathbf{W}}$ as $\tilde{\sigma}_1 \ge \cdots \ge \tilde{\sigma}_P$. Based on the properties of matrix extreme singular values, we have
\begin{eqnarray}
\sigma_1 = \|\mathbf{W}\|_2 = \max_{\mathbf{x}\ne 0}\frac{\|\mathbf{W}\mathbf{x}\|_2}{\|\mathbf{x}\|_2} = \min_{\mathbf{x}\ne 0}\frac{\|\mathbf{W}\mathbf{x}\|_2}{\|\mathbf{x}\|_2} = \sigma_P = 1 . \nonumber
\end{eqnarray}
Let $\mathbf{x}^{*} = \arg\max_{\mathbf{x}\ne 0} \frac{\|\widetilde{\mathbf{W}}\mathbf{x}\|_2}{\|\mathbf{x}\|_2}$, we have
\begin{eqnarray}
\tilde{\sigma}_1 = \frac{\|\widetilde{\mathbf{W}}\mathbf{x}^{*}\|_2}{\|\mathbf{x}^{*}\|_2} = \frac{\|\mathbf{G}\mathbf{W}\mathbf{x}^{*}\|_2}{\|\mathbf{x}^{*}\|_2} \le \frac{\|\mathbf{G}\|_2 \|\mathbf{W}\mathbf{x}^{*}\|_2}{\|\mathbf{x}^{*}\|_2} , \nonumber
\end{eqnarray}
where we have used the fact that $\|\mathbf{A}\mathbf{b}\|_2 \le \|\mathbf{A}\|_2\|\mathbf{b}\|_2$ for any $\mathbf{A} \in \mathbb{R}^{m\times n}$ and $\mathbf{b} \in \mathbb{R}^n$. We thus have
\begin{eqnarray}
\tilde{\sigma}_1 \le \|\mathbf{G}\|_2 \frac{\|\mathbf{W}\mathbf{x}^{*}\|_2}{\|\mathbf{x}^{*}\|_2} \le \|\mathbf{G}\|_2 \max_{\mathbf{x}\ne 0} \frac{\|\mathbf{W}\mathbf{x}\|_2}{\|\mathbf{x}\|_2} = |g_{\max}| . \nonumber
\end{eqnarray}
Since $\mathbf{G}$ has nonzero entries, we have $\mathbf{W} = \mathbf{G}^{-1}\widetilde{\mathbf{G}}$. Let $\mathbf{x}^{*} = \arg\min_{\mathbf{x}\ne 0}\frac{\|\widetilde{\mathbf{W}}\mathbf{x}\|_2}{\|\mathbf{x}\|_2}$, the properties of matrix extreme singular values give $\tilde{\sigma}_P = \frac{\|\widetilde{\mathbf{G}}\mathbf{x}^{*}\|_2}{\|\mathbf{x}^{*}\|_2}$, and $\sigma_P = \min_{\mathbf{x}\ne 0}\frac{\|\mathbf{W}\mathbf{x}\|_2}{\|\mathbf{x}\|_2} = 1$. We thus have
\begin{eqnarray}
1 = \min_{\mathbf{x}\ne 0}\frac{\|\mathbf{G}^{-1}\widetilde{\mathbf{G}}\mathbf{x}\|_2}{\|\mathbf{x}\|_2} \le \frac{\|\mathbf{G}^{-1}\widetilde{\mathbf{G}}\mathbf{x}^{*}\|_2}{\|\mathbf{x}^{*}\|_2} \le \|\mathbf{G}^{-1}\|_2 \frac{\|\widetilde{\mathbf{G}}\mathbf{x}^{*}\|_2}{\|\mathbf{x}^{*}\|_2} , \nonumber
\end{eqnarray}
which gives $\tilde{\sigma}_P \ge |g_{\min}|$. Overall, we have
\begin{eqnarray}
|g_{\max}| \ge \tilde{\sigma}_1 \ge \cdots \ge \tilde{\sigma}_P \ge |g_{\min}| . \nonumber
\end{eqnarray}

We next consider the special case of $M \le N$ and $\textrm{rank}(\mathbf{W}) = M$. Without loss of generality, we assume diagonal entries $\{ g_i \}_{i=1}^M$ of $\mathbf{G}$ are all positive and ordered. By definition we have $\widetilde{\mathbf{W}} = \mathbf{I}\mathbf{G}\mathbf{W}$, where $\mathbf{I}$ is an identity matrix of size $M\times M$. Let $\mathbf{V} = \left[ \mathbf{W}^{\top}, \mathbf{W}^{\bot\top} \right]$, where $\mathbf{W}^{\bot}$ denotes the orthogonal complement of $\mathbf{W}$, we thus have the SVD of $\widetilde{\mathbf{W}}$ by construction as $\widetilde{\mathbf{W}} = \mathbf{I} \left[ \mathbf{G}, \mathbf{0} \right] \mathbf{V}^{\top}$. When some values of $\{ g_i \}_{i=1}^M$ are not positive, the SVD can be constructed by changing the signs of the corresponding columns of either $\mathbf{I}$ or $\mathbf{V}$. Since matrix singular values are uniquely determined (while singular vectors are not), singular values of $\widetilde{\mathbf{W}}$ are thus exactly $\{ |g_i| \}_{i=1}^M$.
\end{proof}

\end{document}